\documentclass[runningheads]{llncs}
\usepackage[T1]{fontenc}
\usepackage{graphicx}
\usepackage{booktabs}
\usepackage{amssymb}
\usepackage{microtype}
\usepackage[misc]{ifsym}

\usepackage{amsmath}
\usepackage{mwe}
\usepackage{tikz}
\usetikzlibrary{positioning, shapes, arrows.meta, decorations.pathreplacing}
\usepackage{enumitem}
\usepackage[colorlinks=true,linkcolor=blue,citecolor=blue,urlcolor=blue]{hyperref}
\usepackage[all]{hypcap}
\usepackage{multirow}
\usepackage{wrapfig}

\begin{document}

\title{GRMLR: Knowledge-Enhanced Small-Data Learning for Deep-Sea Cold Seep Stage Inference}

\titlerunning{Knowledge-enhanced Cold Seep Developmental Stages Inference}
% If the full title of your paper is short enough to also fit in the running head, you can omit the abbreviated paper title here. You can check as follows: if you comment out the \titlerunning line, something will appear in the header of all odd-numbered pages of your PDF from page 3 onward. This something is either the full title (in which case all is well), or the error message "Title Suppressed Due to Excessive Length". If this error message appears, you're going to want to provide an abbreviated title within the \titlerunning command, because if you won't do it, Springer will do it for you.

%N.B.: Author information (both in the \author{} and \authorrunning{} command) should only be present in the Camera-Ready Version of your paper. The version that you initially submit for review, ought to be double-blind. So, when initially submitting your paper, use:
%\author{ }
\author{Chenxu Zhou \thanks{Chenxu Zhou, Zelin Liu and Rui Cai contributed equally to this work.} \and Zelin Liu \protect\footnotemark[1] \and Rui Cai \protect\footnotemark[1] \and Houlin Gong \and Yikang Yu \and Jia Zeng \and Yanru Pei \and Liang Zhang \and Weishu Zhao\textsuperscript{\dag} \and Xiaofeng Gao\textsuperscript{\dag}}

\begingroup
\renewcommand\thefootnote{\fnsymbol{footnote}} % 设置脚注符号为符号形式（*，†，‡等）
\footnotetext[4]{Corresponding authors: zwsh88@sjtu.edu.cn, gao-xf@cs.sjtu.edu.cn.}
\endgroup
% You may leave out the orcidID information, if you want to.
% Use \corr to indicate the corresponding author. Note the spacing around the \corr command. Only one author can be the corresponding author.

%N.B.: comment out the \authorrunning{} command for the double-blind version of your paper submitted for review. Later, if your paper is accepted, use the command for the Camera-Ready Version.

\authorrunning{[Z. Liu] et al.}

% If there are two authors, write 'A.L. Benjamin and C.C. Broadus Jr.'
% If there are more than two authors, '[...] et al.' is used.

\institute{Shanghai Jiao Tong University, 200240 Shanghai, China}
%\institute{ }

\maketitle              % typeset the header of the contribution

\begin{abstract}

% Assessing deep-sea cold seep developmental stages conventionally relies on expensive, high-risk manned submersibles and visual macrofauna surveys. 
Deep-sea cold seep stage assessment has traditionally relied on costly, high-risk manned submersible operations and visual surveys of macrofauna. Although microbial communities provide a promising and more cost-effective alternative, reliable inference remains challenging because the available deep-sea dataset is extremely small ($n=13$) relative to the microbial feature dimension ($p=26$), making purely data-driven models highly prone to overfitting.
% While microbial communities offer a promising cost-effective proxy, constructing reliable inference models is hindered by extremely limited deep-sea samples ($n=13$) relative to high-dimensional microbial features ($p=26$), leading to severe overfitting in purely data-driven approaches. 
To address this, we propose a knowledge-enhanced classification framework that incorporates an ecological knowledge graph as a structural prior. By fusing macro-microbe coupling and microbial co-occurrence patterns, the framework internalizes established ecological logic into a \underline{\textbf{G}}raph-\underline{\textbf{R}}egularized \underline{\textbf{M}}ultinomial \underline{\textbf{L}}ogistic \underline{\textbf{R}}egression (GRMLR) model, effectively constraining the feature space through a manifold penalty to ensure biologically consistent classification. Importantly, the framework removes the need for macrofauna observations at inference time: macro–microbe associations are used only to guide training, whereas prediction relies solely on microbial abundance profiles. Experimental results demonstrate that our approach significantly outperforms standard baselines, highlighting its potential as a robust and scalable framework for deep-sea ecological assessment.

\end{abstract}

\section{Introduction}

\begin{figure}[t]
\centering
\begin{minipage}[t]{0.45\textwidth}
  \centering
  \includegraphics[width=\textwidth]{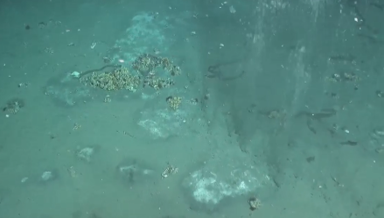}
  \caption{Methane bubbling observed in the cold seep area.}
  \label{fig:methane-eruption}
\end{minipage}
\hfill
\begin{minipage}[t]{0.45\textwidth}
  \centering
  \includegraphics[width=\textwidth]{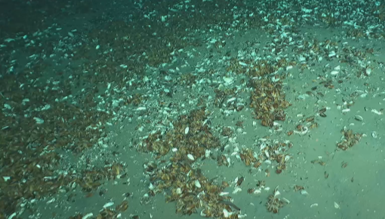}
  \caption{Rich biological assemblages around a cold seep.}
  \label{fig:rich-fauna}
\end{minipage}
\end{figure}

\begin{figure}[t]
    \centering
    \includegraphics[width=0.95\textwidth]{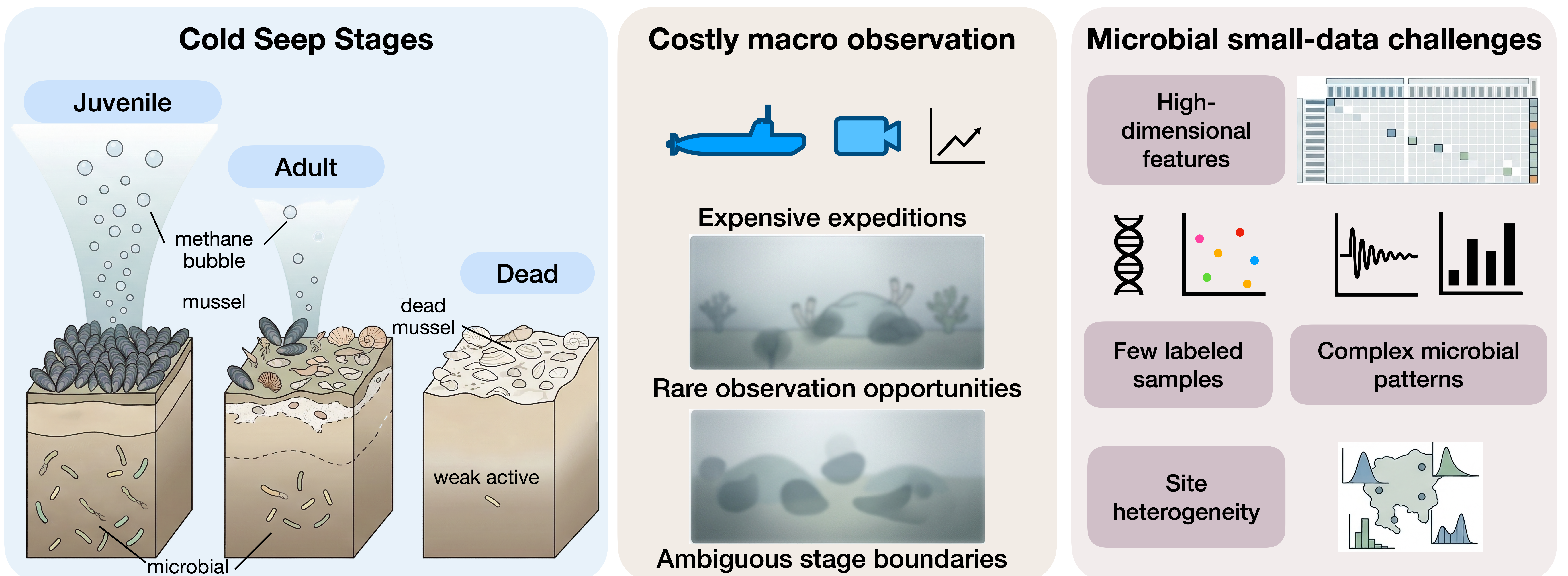}
    \caption{\textbf{Motivation:} Cold seep development stage recognition requires robust learning from complex microbial data, since macrofaunal observations are expensive and challenging to acquire.}
    \label{fig:moti}
\end{figure}

Deep-sea cold seeps are methane-fueled chemosynthetic ecosystems characterized by tight coupling between seepage-driven macrofaunal assemblages and methane-cycling microbial communities~\cite{levin2005ecology}. Beyond their ecological uniqueness, they are important natural systems for probing life in extreme environments, sustaining deep-sea biodiversity hotspots, and regulating marine methane turnover and carbon burial~\cite{akam2023methane}. Cold seeps further exhibit stage-like dynamics, typically transitioning from juvenile to adult and dead states, as illustrated in Fig.~\ref{fig:moti}, with coordinated shifts in seepage, geochemistry, microbes, and fauna~\cite{levin2005ecology}. Representative field observations of methane bubbling and rich cold seep biota are shown in Fig.~\ref{fig:methane-eruption} and Fig.~\ref{fig:rich-fauna}. Consequently, stage recognition is a prerequisite for resolving seep evolution and for evaluating roles in methane filtration, carbon sequestration, and ecosystem vulnerability~\cite{akam2023methane}.

Despite growing interest in cold seep stage recognition, existing assessment paradigms remain bottlenecked by the cost and sparsity of observation (Fig.~\ref{fig:moti}). Current practice still relies largely on video-based surveys of macrofaunal assemblages collected by ship-supported deep-sea platforms, yet these observations are expensive, geographically biased, and difficult to scale over space and time~\cite{bell2025little}. The problem is further complicated by the fact that seep habitats typically form patchy mosaics with broad transition zones, so stage boundaries are often visually blurred rather than explicitly observable~\cite{levin2005ecology}. This motivates a microbial route: compared with sparse visual snapshots, microbial communities are more tightly coupled to methane-driven geochemical conditions and therefore offer a potentially more direct signal of seep activity~\cite{niu2023methane}. However, this alternative introduces a new inference challenge, because the available microbial data are small-sample, high-dimensional, and compositional, while the ecological mechanism that maps microbial variation to seep developmental stages remains only partially understood \cite{cawley2010over}. As a result, cold seep stage recognition requires more than replacing images with omics features; it calls for a learning framework that is robust to small-data statistics and constrained by ecological structure \cite{levin2005ecology}.

To address these challenges, we propose a \underline{\textbf{G}}raph-\underline{\textbf{R}}egularized \underline{\textbf{M}}ultinomial \underline{\textbf{L}}ogistic \underline{\textbf{R}}egression (GRMLR) model that improves both the information efficiency and biological validity of limited microbial data through two tightly coupled designs. First, we apply the centered log-ratio transformation to convert compositional microbial abundances from the simplex into a Euclidean space, enabling stable optimization and ensuring that the learned patterns reflect genuine taxonomic variation rather than artifacts induced by relative-abundance constraints \cite{gloor2017microbiome}. Building on this foundation, we further construct an Ecological Knowledge Graph that encodes expert-informed microbe–macrofauna coupling as a graph-structured prior, thereby injecting missing ecological logic into the learning process \cite{suess_marine_2014,belkin2006manifold,zhou2026knowledge}. Rather than compensating for data scarcity with more samples, our framework leverages structured prior knowledge to bridge microbial signatures and macro-ecological stages, allowing the model to internalize discriminative ecological cues that are otherwise accessible only through costly visual surveys. As a result, the proposed framework turns sparse microbial observations into a robust proxy for cold seep stage recognition, enabling accurate inference without relying on expensive macro-level exploration. The main contributions of this work are summarized as follows:
\begin{itemize}[leftmargin=*]
    \item \textbf{New problem formulation.}
    We formulate cold seep developmental stage recognition as a microbial-driven small-data classification problem, offering a scalable alternative to expensive macrofauna-based visual assessment.

    \item \textbf{Knowledge-enhanced modeling.}
    We propose a graph-regularized framework that injects an Ecological Knowledge Graph into multinomial logistic regression, allowing macro-microbe coupling and microbial co-occurrence structure to guide biologically consistent classification under extreme data scarcity.

    \item \textbf{Decoupled deployment mechanism.}
    Our framework uses macrofauna know-ledge only during training, while requiring only microbial features at inference time, thus decoupling deployment from macro-level exploration.

    \item \textbf{Strong empirical performance.}
    Experiments demonstrate clear improvements over standard baselines, validating the robustness and practical value of the proposed framework for deep-sea ecological assessment.
\end{itemize}

\section{Preliminary} \label{sec:preli}

\begin{figure}[t]
    \centering
    \includegraphics[width=0.95\textwidth]{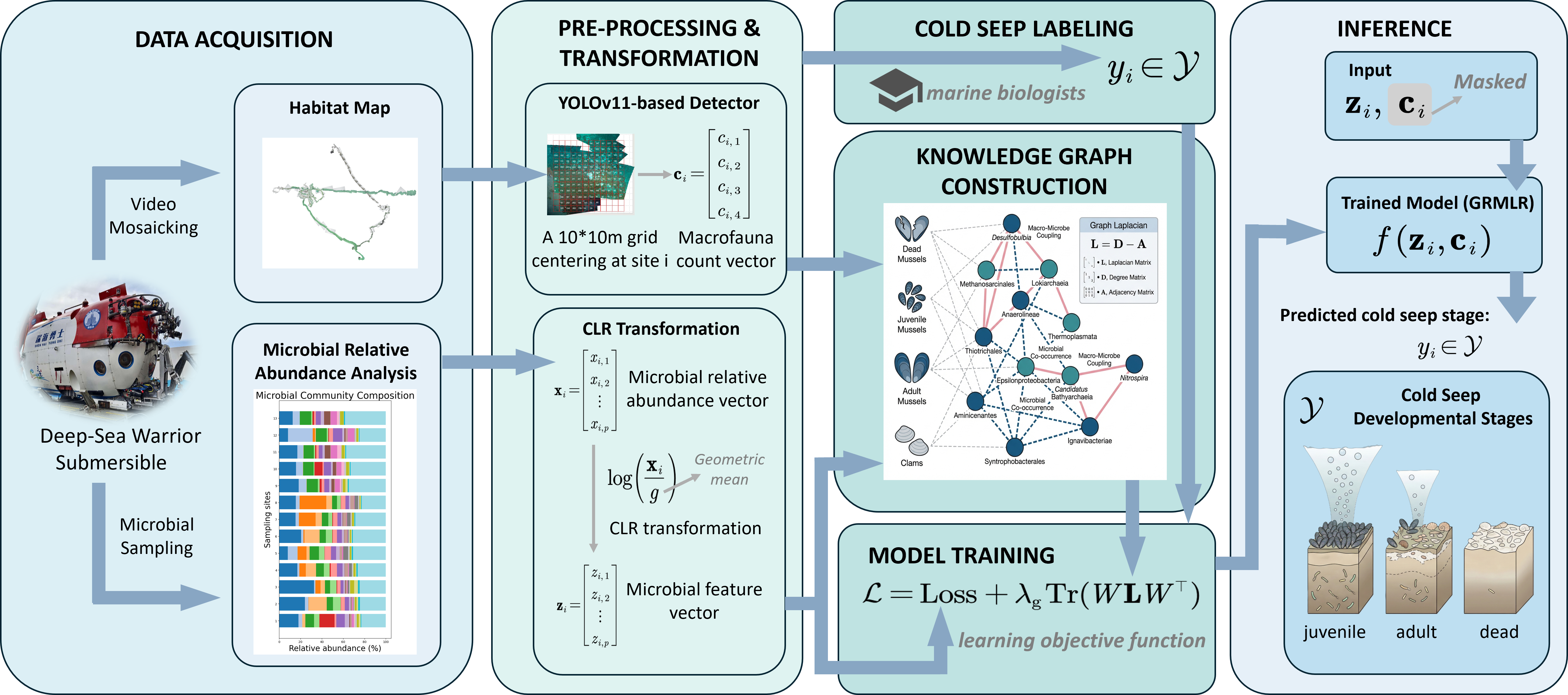}
    \caption{\textbf{Framework of GRMLR and data flow.} Macrofauna counts and CLR-transformed microbial features are integrated through an ecological knowledge graph during training, while inference uses only microbial data for stage classification.}
    \label{fig:framework}
\end{figure}

% Due to the high risks and excessive costs associated with acquiring deep-sea imagery and macrofauna data, this study develops a cold seep developmental stages classification framework integrating a microbe-macrofauna Knowledge Graph. \\

\noindent \textbf{Data Representation and Compositional Constraints}
The experimental data are collected from $n=13$ deep-sea cold seep sampling sites, indexed by $i \in \{1, \dots, n\}$. For each site $i$, we define three key biological observation vectors: 

(1) A $p$-dimensional microbial relative abundance vector is defined as $\mathbf{x}_i = \langle x_{i,1}, x_{i,2}, \dots, x_{i,p} \rangle \in \Delta^{p-1}$, where $p=26$ is the number of taxonomic classes. Each component $x_{i,j}$ represents the observed relative abundance of microbial class $j$ at site $i$, satisfying the compositional sum-to-one constraint:
\begin{equation}
x_{i,j} > 0 \quad (\forall j), \quad \sum_{j=1}^p x_{i,j} = 1
\end{equation}
This constraint restricts the raw microbial relative abundance vectors $\mathbf{x}_i$ to reside on a $(p-1)$-dimensional probability simplex $\Delta^{p-1}$. In such a closed geometry, standard Euclidean matrix operations and statistics (e.g., correlations and covariance) are distorted by spurious dependencies, and direct linear modeling is hampered by multicollinearity, making parameter estimation unreliable. More importantly, it imposes spurious correlations between species, which can obscure true biological signals.\cite{aitchison1982statistical}

(2) A $k$-dimensional macrofauna count vector is defined as $\mathbf{c}_i = \langle c_{i,1}, c_{i,2}, \dots, \\ c_{i,k} \rangle \in \mathbb{R}^k$, where $k=4$. These counts quantify key macrofauna categories (dead, adult, juvenile mussels, and clams) identified via computer vision from spatially aligned habitat maps. This vector $\mathbf{c}_i$ serves as quantitative biological evidence of the cold seep developmental stages.

(3) Each site is annotated with a categorical label $$y_i \in \mathcal{Y} = \{ \text{juvenile, adult, dead} \},$$ representing its typical ecological developmental stage determined by experts based on the habitat configuration.

To overcome the compositional artifacts inherent in microbial relative abundance data, we map the constrained simplex $\Delta^{p-1}$ into a tractable Euclidean feature space $\mathbb{R}^p$ using the Centered Log-Ratio (CLR) transformation \cite{gloor2017microbiome}. The transformed microbial feature vector $\mathbf{z}_i = \langle z_{i,1}, \dots, z_{i,p} \rangle$ is computed as:
\begin{equation}
z_{i,j} = \text{CLR}(\mathbf{x}_i)_j = \log(x_{i,j} + \epsilon) - \frac{1}{p} \sum_{k=1}^p \log(x_{i,k} + \epsilon)
\end{equation}
where $\epsilon = 10^{-6}$ is a pseudo-count to handle zero values.\\

\noindent \textbf{Ecological Knowledge Graph and Laplacian}
To incorporate expert-informed ecological priors into the data-scarce classification problem, we introduce an Ecological Knowledge Graph, denoted as $\mathcal{G} = (\mathcal{V}, \mathcal{E})$.

The vertex set $\mathcal{V} = \{v_1, v_2, \dots, v_p\}$ represents the $p=26$ microbial taxonomic classes. Let $\mathbf{A} \in \mathbb{R}^{p \times p}$ be the weighted adjacency matrix of $\mathcal{G}$, where the edge weight $A_{uv} \geq 0$ quantifies the strength of ecological association between microbial taxa $u$ and $v$, derived from dual biological sources: macrofauna-induced taxa similarity and microbial co-occurrence structures. $\mathbf{A}$ is required to be symmetric.

The structural information encoded in the graph $\mathcal{G}$ can be transformed into a manifold regularization prior using the combinatorial Graph Laplacian matrix $\mathbf{L}$. The Graph Laplacian $\mathbf{L} \in \mathbb{R}^{p \times p}$ is defined as \cite{belkin2006manifold}:
\begin{equation}
\label{eq:laplac}
\mathbf{L} = \mathbf{D} - \mathbf{A}
\end{equation}
where $\mathbf{D} \in \mathbb{R}^{p \times p}$ is the diagonal degree matrix with elements $D_{uu} = \sum_{v \in \mathcal{V}} A_{uv}$. This Laplacian matrix $\mathbf{L}$ is positive semi-definite and acts as a smoother over the microbial taxa space.\\

\noindent \textbf{Task Formalization}
The final objective is to learn a supervised classification mapping function $f: \mathbb{R}^p \times \mathbb{R}^k \times \mathcal{G} \to \mathcal{Y}$. This function predicts the cold seep developmental stage label $y_i$ for site $i$ from the input data.

% \section{Method}
\section{Knowledge-Enhanced Cold Seep Stage Classificatio}

We implement an end-to-end workflow from "Deep-Sea Warrior" manned submersible video to cold seep developmental stage classification (Fig.~\ref{fig:framework}). The pipeline has four steps: (i) constructing continuous 2D habitat maps $\mathcal{M}$ by image mosaicking, then aligning the 13 sampling sites to extract $10 \times 10$ m grids and using a YOLO-based detector to quantify macrofauna and generate count vectors $\mathbf{c}_i$ (Section~\ref{sec:habitat-macro}); (ii) defining stage labels $y_i \in \mathcal{Y}$ through expert annotation based on macrofauna count vectors and habitat maps; (iii) analyzing microbial relative abundances $\mathbf{x}_i$ and generating microbial feature vectors $\mathbf{z}_i$ with the Centered Log-Ratio (CLR) transformation (Section~\ref{sec:micro}); and (iv) training a knowledge-graph-regularized classifier to predict stage labels $y_i$, so inference is decoupled from macrofauna count vectors $\mathbf{c}_i$ (Section~\ref{sec:model}).

\subsection{Habitat Map Construction and Macrofauna Detection} \label{sec:habitat-macro}

\begin{figure}[t]
\centering
\begin{minipage}[t]{0.45\textwidth}
  \centering
  \includegraphics[width=\textwidth]{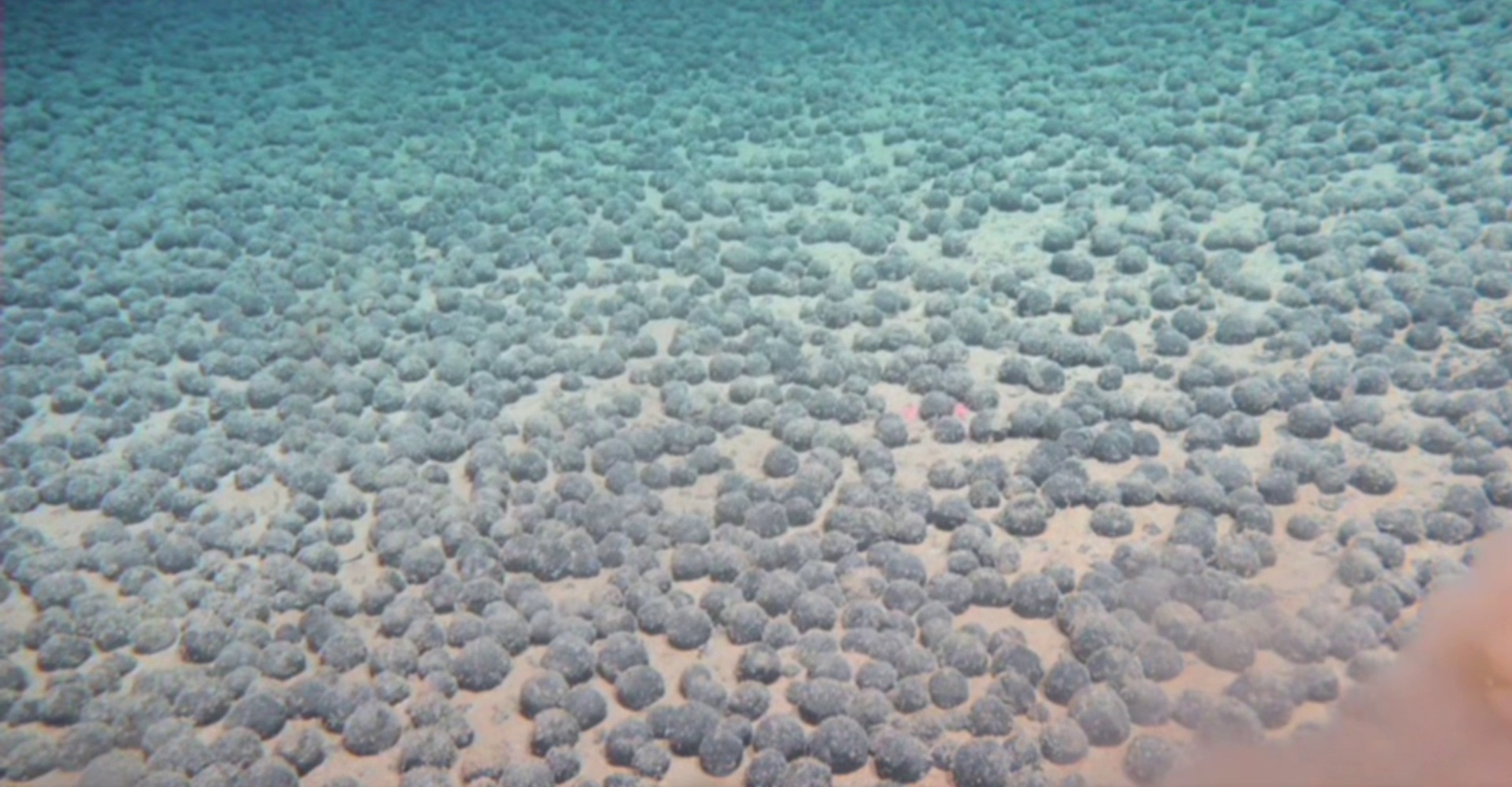}
  \caption{A representative raw submersible frame ($1920\times1080$) used as input to the mapping pipeline.}
  \label{fig:raw-frame}
\end{minipage}
\hfill
\begin{minipage}[t]{0.45\textwidth}
  \centering
  \includegraphics[width=\textwidth]{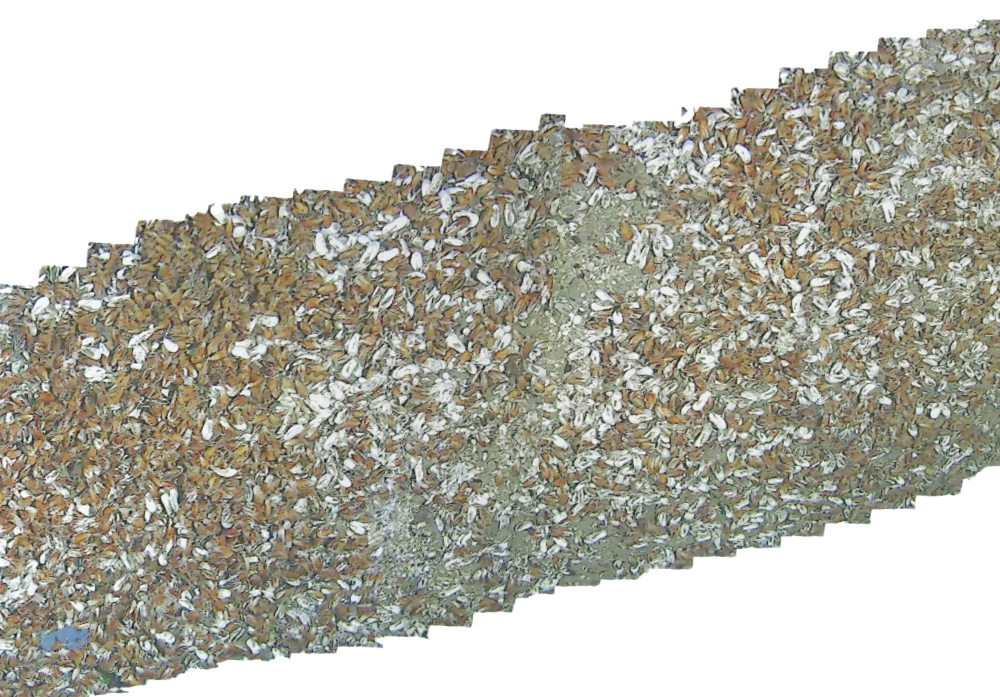}
  \caption{A stitched habitat map tile produced by DUSt3R, showing a cold seep region with dense macrofauna.}
  \label{fig:map-tile}
\end{minipage}
\end{figure}

Manned submersible footage suffers from inconsistent viewing angles, varying camera-to-seafloor distances, frequent heading changes, and unavailable camera intrinsics, causing classical structure-from-motion pipelines to produce tearing artefacts, large gaps, or erroneous overlaps. We therefore adopt DUSt3R \cite{wang2024dust3r}, which recovers dense 3D geometry from uncalibrated image pairs.

Each raw frame $I_t$ undergoes colour correction and contrast enhancement (Fig.~\ref{fig:raw-frame}). Each $1920\times1080$ frame is first vertically cropped to the central $1920\times540$ band (rows 271--810), discarding the featureless top and bottom margins. The resulting strip is then divided into three overlapping $768\times540$ sub-images---left $S_t^L$ (columns 1--768), centre $S_t^C$ (577--1344), and right $S_t^R$ (1153--1920)---yielding a 191-pixel lateral overlap between adjacent patches to preserve high image resolution within each strip and to ensure sufficient shared content for cross-frame correspondence. These strips are passed into DUSt3R, which jointly estimates 3D pointmaps and camera poses, yielding a globally consistent point cloud. The cloud is projected onto a plane perpendicular to the seafloor normal to produce a top-down habitat map, and multiple local maps from different sessions are merged hierarchically into the final $\mathcal{M}$. Frames are sampled at one per 30 frames ($\approx$1 s), yielding roughly 24{,}000 frames per map; the resulting maps cover areas ranging from $100\,\text{m}\times150\,\text{m}$ to $450\,\text{m}\times480\,\text{m}$.

\noindent We then perform object detection on $\mathcal{M}$ to quantify the ecological indicators of cold seep development; Fig.~\ref{fig:map-tile} shows a representative stitched map tile used for this stage. Working on the stitched map rather than raw video frames avoids duplicate counting from overlapping views.

\textbf{Dataset Construction and Annotation.}
We build a macrofauna dataset by extracting high-resolution patches from $\mathcal{M}$, which are manually annotated by marine biologists into four categories: adult, juvenile, and dead individuals, and \emph{Calyptogena} clams. Let the annotated dataset be $\mathcal{D}_{\text{ann}} = \{(\mathcal{P}_k, \mathcal{B}_k)\}_{k=1}^{N}$, where $\mathcal{P}_k$ is an image patch and $\mathcal{B}_k$ the corresponding bounding boxes with class labels. Random flipping, rotation, and colour jittering are applied as data augmentation to improve robustness against varying lighting and seafloor sediment coverage.

\textbf{Model Fine-tuning with YOLOv11.}
We adopt YOLOv11 for its strong balance of speed and accuracy, following its broad use in underwater organism detection \cite{zhao_mdg_yolo_2025}. The model is initialised with pretrained weights and then fine-tuned on $\mathcal{D}_{\text{ann}}$ via transfer learning, with a combined localisation, objectness, and classification loss optimised by stochastic gradient descent.

\textbf{Map-based Inference and Aggregation.}
For detection, $\mathcal{M}$ is split into overlapping tiles $\{\mathcal{T}_j\}$ to match the network's input size. Each tile is processed by the fine-tuned detector, and predictions are mapped back into the global coordinate system of $\mathcal{M}$. Non-Maximum Suppression (NMS) is then applied across all merged predictions to remove duplicates \cite{zhao_mdg_yolo_2025}. The final counts are aggregated into the macrofauna count vector: $\mathbf{c}_i=\langle c_{i,1},c_{i,2},c_{i,3},c_{i,4} \rangle$.

\subsection{Microbial Relative Abundance Analysis} \label{sec:micro}

\noindent \textbf{Metagenomic Sequencing and MAG Assembly} 
Sediment samples were collected from 13 sampling sites using pressure-retaining containers and preserved at -80°C. We extracted genomic DNA from these samples and performed metagenomic sequencing on the DNBSEQ-T1 platform (BGI Genomics Co., Ltd., Shenzhen, China) in paired-end 150 bp mode. To quantify the microbial community, the raw reads were assembled into contigs, which were further clustered into Metagenome-Assembled Genomes (MAGs) using a multi-tool binning and refinement pipeline \cite{li2009bwa}. These MAGs serve as the representative genomic units for subsequent taxonomic and abundance analysis.

\noindent \textbf{Taxonomic Assignment and Vector Construction} 
We performed taxonomic assignment for the recovered MAGs using the GTDB database. We used GTDB-Tk and related GTDB resources for taxonomy annotation \cite{chaumeil2022gtdbtk_v2,parks2020complete}. To construct the microbial relative abundance vector $\mathbf{x}_i$, we first computed the coverage depth of each MAG per sample. We then aggregated these MAGs at the class level and categorized them into 26 distinct groups: the 25 most abundant microbial classes across all samples, and one aggregate category for all remaining classes. For each sampling site $i$, the resulting counts were normalized to represent the relative abundance of each class, forming the $p$-dimensional vector $\mathbf{x}_i = \langle x_{i,1}, \dots, x_{i,p} \rangle$ (where $p=26$).

\noindent \textbf{CLR Transformation} 
As analyzed in the Section~\ref{sec:preli}, the raw abundance vector $\mathbf{x}_i$ is subject to compositional constraints that introduce multicollinearity. To map the data into a more discriminative and unconstrained feature space, we applied the Centered Log-Ratio (CLR) transformation to $\mathbf{x}_i$. This process yields the final microbial feature vector $\mathbf{z}_i$.

\subsection{Graph-Regularized Multinomial Logistic Regression} \label{sec:model}

This section details the proposed Graph-Regularized Multinomial Logistic Regression (GRMLR) framework. By integrating the Ecological Knowledge Graph $\mathcal{G}$ defined in Section~3, the model targets the small-sample classification challenge ($n=13$, $p=26$). The framework utilizes expert-informed logic to bridge the gap between low-level microbial signatures ($\mathbf{z}_i$) and high-level ecological states ($y_i$), enabling robust classification during inference using only microbial data.

\noindent \textbf{Adjacency Matrix Construction}
To operationalize the Ecological Knowledge Graph $\mathcal{G}$ and internalize the macro-microbe association into the classifier \cite{zhou2026knowledge}, we construct its weighted adjacency matrix $\mathbf{A} \in \mathbb{R}^{p \times p}$ by fusing two biological dimensions, thereby encoding ecological dependencies as structural priors:

\noindent\textbf{Macro-Microbe Coupling ($\mathbf{A}_{\text{macro}}$)}: Edge weights are derived from shared Spearman correlation patterns between microbial taxa and macrofauna count vectors ($\mathbf{c}_i$). Specifically, for any two microbial taxa $u, v \in \mathcal{V}$, their similarity is calculated based on how closely their respective abundances correlate with the counts of dead mussels, adult mussels, juvenile mussels, and clams across training samples. This captures the indirect ecological dependencies between microbial communities and cold seep stages as reflected by macrofauna distributions.

\noindent\textbf{Microbial Co-occurrence ($\mathbf{A}_{\text{co}}$)}: Edges are formed based on pairwise Spearman correlations between microbial feature ($\mathbf{z}_i$) across all sampling sites. This component represents intrinsic symbiotic relationships or shared ecological niches within the microbial community in response to methane availability.

The final symmetric adjacency matrix $\mathbf{A}$ is obtained by a weighted fusion of these two sources: $\mathbf{A} = \alpha \mathbf{A}_{\text{macro}} + (1-\alpha) \mathbf{A}_{\text{co}}$, where $\alpha \in [0,1]$ is a hyperparameter balancing the two ecological dimensions. The corresponding Graph Laplacian $\mathbf{L}$ is then computed as formalised in Equation~\ref{eq:laplac}, internalising these macro-microbe associations for use in the classifier \cite{belkin2006manifold,suess_marine_2014}.

\noindent \textbf{Graph-Regularized Multinomial Logistic Regression}
We propose the Gra-ph-Regularized Multinomial Logistic Regression (GRMLR) framework to predict the cold seep developmental stage $y_i \in \mathcal{Y}$ from the CLR-transformed microbial feature vector $\mathbf{z}_i \in \mathbb{R}^p$. Let $W \in \mathbb{R}^{K \times p}$ be the weight matrix and $\mathbf{b} \in \mathbb{R}^K$ be the bias vector, where $K=3$ is the number of stages and $p=26$ is the number of microbial classes. The model predicts the probability of sample $\mathbf{z}_i$ belonging to class $k$ using the softmax function:
\begin{equation}
P(y_i = k \mid \mathbf{z}_i) = \frac{\exp(\mathbf{w}_k^\top \mathbf{z}_i + b_k)}{\sum_{j=1}^K \exp(\mathbf{w}_j^\top \mathbf{z}_i + b_j)}
\end{equation}
The optimal parameters are learned by minimizing the following loss function:
\begin{equation}
\label{eq:loss}
\mathcal{L}(W, \mathbf{b}) = \underbrace{\frac{1}{n} \sum_{i=1}^{n} -\log P(y_i \mid \mathbf{z}_i)}_{\text{Cross-Entropy Loss}} + \underbrace{\lambda_{\text{l2}} \|W\|_F^2}_{\ell_2 \text{ Regularization}} + \underbrace{\lambda_{\text{g}} \mathrm{Tr}(W \mathbf{L} W^\top)}_{\text{Graph Regularization}}
\end{equation}
The component is the Graph Laplacian penalty, $\mathrm{Tr}(W \mathbf{L} W^\top)$, which can be rewritten as $\frac{1}{2}\sum_{u,v} A_{uv} \| \mathbf{w}_{:,u} - \mathbf{w}_{:,v} \|^2$ \cite{belkin2006manifold}. This term acts as a manifold constraint, forcing ecologically related taxa (those with high similarity $A_{uv}$ in the KG) to have similar weight vectors in the classifier. By injecting ecological logic into the optimization, this penalty prevents overfitting to noisy features in small samples and ensures that the classification results are biologically consistent.

\noindent \textbf{Training and Inference Stages}
However, as formalised in the end of Section~\ref{sec:preli}, a critical constraint of this study is that the macrofauna data is not available at the inference time. This leads to distinct model inputs in different operational phases, and our knowledge-enhanced framework explicitly decouples these modes to utilise all available data during training while allowing microbial-only assessment during deployment:

\emph{Training Stage}: The framework is provided with CLR-transformed microbial feature vectors $\{\mathbf{z}_i\}$, macrofauna count vectors $\{\mathbf{c}_i\}$, and ground-truth labels $\{y_i\}$. It utilizes the fusion of all three signals ($\mathbf{z}_i$ and $\mathbf{c}_i$ for labels $y_i$). The macrofauna count vectors are used \textit{only} to construct the KG topology (specifically $\mathbf{A}_{\text{macro}}$). The framework then learns weights $W$ and bias $\mathbf{b}$ by minimizing the graph-regularized loss function in Equation~\ref{eq:loss}. This process forces the classifier to internalize the macrofauna ecological logic into its parameter weights.

\emph{Inference Stage}: The model is decoupled from the macrofauna count vectors. It performs classification on new samples using \textit{only} the CLR-transformed microbial feature vector $\mathbf{z}_i$ and the pre-optimized Knowledge Graph parameters ($W, \mathbf{b}$). In this way, the model leverages the internalized macrofauna priors from the training stage to achieve accurate cold seep assessment even when expensive deep-submergence imaging and macrofauna surveys are not available.

\section{Experiments}

We evaluate the proposed framework on a real-world deep-sea cold seep dataset to address the following research questions (RQ):

\noindent \textbf{RQ1: Effectiveness.} How does GRMLR perform compared with standard compositional baselines and established machine-learning methods?

\noindent \textbf{RQ2: Modularity.} How do the different components---CLR transformation, macro-induced adjacency $\mathbf{A}_{\mathrm{macro}}$, co-occurrence adjacency $\mathbf{A}_{\mathrm{co}}$, and graph regularisation itself---contribute to overall performance?

\noindent \textbf{RQ3: Sensitivity.} How does the graph mixing parameter $\alpha$ affect classification accuracy, and how broad is the optimal operating range?

\noindent \textbf{RQ4: Interpretability.} Which microbial taxa are the most informative predictors of cold seep developmental stages, and do the identified taxa align with established ecological knowledge?

\subsection{Experimental Environment}

\noindent\textbf{Dataset.}
The evaluation is conducted on aligned deep-sea sampling sites collected from the South China Sea cold seep field \cite{feng_cold_2018}. The dataset comprises $n=13$ sites, each characterised by a microbial relative abundance vector over $p=26$ taxonomic classes and an auxiliary macrofauna count vector over four categories (dead, adult, juvenile, and \emph{Calyptogena} clam). Sites are labelled into three developmental stages by marine biologists: \emph{Juvenile} ($n_{\text{J}}=3$), \emph{Adult} ($n_{\text{A}}=7$), and \emph{Dead} ($n_{\text{D}}=3$). Table~\ref{tab:dataset} summarises the dataset statistics.

\begin{table}[t]
\centering
\caption{Dataset statistics of the cold seep sampling sites.}
\label{tab:dataset}
\begin{tabular}{ll}
\toprule
Statistic & Value \\
\midrule
No.\ of sampling sites $n$ & 13 \\
No.\ of microbial taxa $p$ & 26 \\
No.\ of developmental stages $K$ & 3 \\
Stage distribution & Juvenile $n_J{=}3$ / Adult $n_A{=}7$ / Dead $n_D{=}3$ \\
No.\ of macrofauna categories & 4 (dead, adult, juvenile, \emph{Calyptogena}) \\
\midrule
\multicolumn{2}{l}{\textit{Microbial taxa ($p=26$, class-level):}} \\[2pt]
\multicolumn{2}{l}{%
\begin{tabular}[t]{@{}lll@{}}
\emph{Gammaproteobacteria} & {DSM-4660}            & \emph{Dissulfuribacteria} \\
\emph{Desulfobacteria}     & \emph{Paceibacteria}       & \emph{Lokiarchaeia} \\
\emph{Dehalococcoidia}     & \emph{Aminicenantia}       & \emph{Gemmatimonadetes} \\
\emph{Syntropharchaeia}    & \emph{Nanoarchaeia}        & \emph{Campylobacteria} \\
\emph{Methanosarcinia}     & \emph{Alphaproteobacteria} & \emph{Desulfobulbia} \\
\emph{Anaerolineae}        & {JS1}                 & \emph{Krumholzibacteria} \\
\emph{Phycisphaerae}       & \emph{Brocadiia}           & \emph{Humimicrobiia} \\
\emph{Bacteroidia}         & \emph{Bathyarchaeia}       & Others \\
\emph{Microgenomatia}      & \emph{Nitrososphaeria}     & \\
\end{tabular}} \\
\bottomrule
\end{tabular}
\end{table}

\noindent\textbf{Evaluation Protocol.}
We adopt leave-one-out cross validation (LOOCV) \cite{wong_performance_2015} as the primary evaluation protocol to maximise training-set utilisation. We report two metrics: {Accuracy} and {Macro-F1}, the latter being particularly relevant under class imbalance.

\noindent\textbf{Permutation Tests.}
The null hypothesis $H_0$ states that the observed accuracy can be obtained by chance. For GRMLR, 50 label permutations are performed. We consider $p < 0.05$ to indicate statistical significance.

\noindent\textbf{Baselines.}
We compare GRMLR against seven baselines: {LR (CLR$+$L2)}, {SVM-RBF (CLR)} \cite{cortes1995support}, {Random Forest (CLR)} \cite{breiman2001random}, {KNN $k{=}3$ (CLR)} \cite{cover1967nearest}, and {LR (Raw)} (without CLR transformation, to isolate the effect of compositional correction). We also include two LLM baselines using Gemini~3~Flash: a {zero-shot} variant that receives only the microbial profile of the test site, and a {LOOCV} variant that additionally receives 12 labelled in-context examples from the training fold.

\noindent\textbf{Implementation Details.}
The GRMLR hyperparameters are selected via exhaustive grid search over 2\,112 configurations with LOOCV accuracy as the selection criterion. The final configuration is: CLR pseudo-count $\epsilon=10^{-6}$, macro-correlation threshold $\tau=0.7$, co-occurrence threshold $\gamma=0.9$, graph mixing $\alpha=0.1$, $\ell_2$ penalty $\lambda_{\ell_2}=0.02$, and graph penalty $\lambda_g=5.0$. Optimisation is carried out via L-BFGS-B \cite{byrd1995limited} with convergence tolerances $\texttt{ftol}=10^{-14}$ and $\texttt{gtol}=10^{-9}$. Class-balanced sample weights are applied to handle the uneven class distribution. The grid search is executed on an Intel Core i9-13900HX CPU using 8 performance cores in parallel, completing in approximately 12 core-hours.

\subsection{Performance Comparison (RQ1)}

\begin{table}[t]
\centering
\caption{LOOCV classification results on the cold seep dataset. The last three columns report the number of correctly predicted samples for each developmental stage (Juvenile: $n_J{=}3$; Adult: $n_A{=}7$; Dead: $n_D{=}3$). The best results are shown in \textbf{Bold}, while the second-best results are \underline{Underlined}.}
\label{tab:main}
\begin{tabular}{lcc ccc}
\toprule
\multirow{2}{*}{Method} & \multirow{2}{*}{Accuracy} & \multirow{2}{*}{Macro-F1} & \multicolumn{3}{c}{Stage-wise correct} \\
\cmidrule(lr){4-6}
 & & & Juvenile & Adult & Dead \\
\midrule
LR (CLR$+$L2) & 0.6154 & \underline{0.6000} & {2} & {5} & 1 \\
SVM-RBF (CLR) & 0.6154 & 0.5794 & 1 & {5} & 2 \\
Random Forest (CLR) & 0.4615 & 0.3072 & 0 & {5} & 1 \\
KNN $k{=}3$ (CLR) & 0.3846 & 0.1852 & 0 & {5} & 0 \\
LR (Raw) & 0.4615 & 0.4222 & 1 & 4 & 1 \\
\midrule
Gemini 3 Flash (zero-shot) & 0.3846 & N/A & 1 & 4 & 0 \\
Gemini 3 Flash (LOOCV) & \underline{0.6923} & N/A & 1 & {5} & {3} \\
\midrule
\textbf{GRMLR (Ours)} & \textbf{0.8462} & \textbf{0.8250} & {2} & {7} & {2} \\
\bottomrule
\end{tabular}
\end{table}

\begin{figure}[t]
\centering
\includegraphics[width=\textwidth]{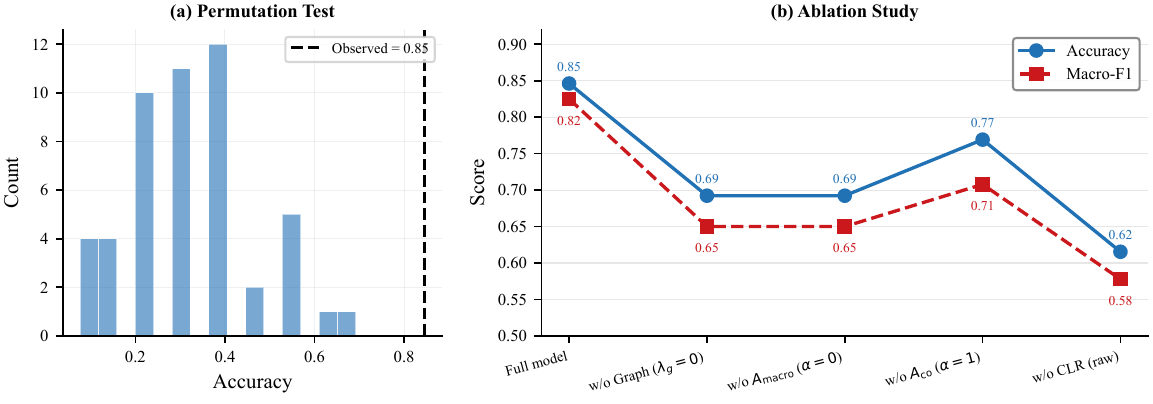}
\caption{(a)~Permutation test null distribution for GRMLR ($p=0.02$); the dashed line marks the observed accuracy of 84.62\%. (b)~Ablation study results; each point shows accuracy and Macro-F1 when the corresponding component is removed.}
\label{fig:permutation}\label{fig:ablation}
\end{figure}

The results in Table~\ref{tab:main} and Figure~\ref{fig:permutation}(a) show that \textbf{GRMLR outperforms all baselines.}
GRMLR attains 84.62\% accuracy and 0.825 Macro-F1 ($p=0.02$), ahead of every baseline by at least 15 points. Stage-wise, all baselines plateau at 5/7 on the Adult majority class and collapse on the two minority stages (0--2 correct each). GRMLR is the only method to achieve a perfect Adult score (7/7) while maintaining 2/3 on both minority stages, showing that graph regularisation lifts all classes uniformly. 
This pattern is consistent with the goal of the model design. In this task, the main difficulty is not fitting the majority class, but learning stable boundaries for the two minority stages from only 13 samples. GRMLR improves exactly at this point: instead of relying on isolated taxa, it uses the ecological graph to encourage coordinated signals from related taxa, which makes the decision boundary more stable under severe data scarcity.
In-context examples help Gemini but cannot close the gap.
Zero-shot Gemini~3~Flash scores only 38.46\% (J:1, A:4, D:0), missing the Dead stage entirely. Adding 12 in-context examples (LOOCV) recovers Dead (3/3) and raises accuracy to 69.23\%, yet Juvenile remains at 1/3---indicating that verbal descriptions of microbial abundances cannot resolve subtle Juvenile--Adult boundaries that the ecological graph prior encodes directly.
CLR is necessary but not sufficient. LR~(Raw) (46.15\%) vs.\ LR~(CLR$+$L2) (61.54\%) shows a 15-point gain from correcting compositional closure alone; however, accuracy stalls 23 points below GRMLR without the graph term.

\subsection{Ablation Study (RQ2)}

To assess the contribution of each component, we compare the following GRMLR variants under identical hyperparameters: 
\textbf{w/o Graph} ($\lambda_g=0$): removes graph regularisation entirely. 
\textbf{w/o $\mathbf{A}_{\mathrm{macro}}$} ($\alpha=0$): retains only microbial co-occurrence structure. 
\textbf{w/o $\mathbf{A}_{\mathrm{co}}$} ($\alpha=1$): retains only macro-induced taxa similarity. 
\textbf{w/o CLR}: feeds raw relative abundances instead of CLR-transformed features. 
Results are presented in Figure~\ref{fig:ablation}(b).

\noindent \textbf{Graph regularisation is the most important component.} Removing it causes a 15.4-point accuracy drop (84.62\% $\to$ 69.23\%), highlighting the central role of the Ecological Knowledge Graph in the small-sample regime.

\noindent \textbf{Both adjacency components are necessary.}
Removing $\mathbf{A}_{\mathrm{macro}}$ alone ($\alpha=0$) produces the same 15.4-point drop as removing the full graph, showing that macro-induced similarity carries the primary structural signal. Removing $\mathbf{A}_{\mathrm{co}}$ ($\alpha=1$) gives a smaller 7.7-point loss (84.62\% $\to$ 76.92\%), indicating that co-occurrence edges provide useful complementary information. The full model outperforms either source in isolation, confirming the value of the dual-source design.
This result also clarifies the logic of graph fusion: $\mathbf{A}_{\mathrm{macro}}$ provides stage-related ecological guidance, while $\mathbf{A}_{\mathrm{co}}$ supplies additional neighbourhood structure among taxa. Their combination therefore improves not only accuracy, but also the biological consistency of the learned classifier.

\noindent \textbf{CLR preprocessing is indispensable.}
Replacing CLR with raw relative abundances gives the biggest single-component drop (84.62\% $\to$ 61.54\%), confirming that handling compositional closure is essential for reliable modelling.

\subsection{Sensitivity Analysis (RQ3)}

The graph mixing parameter $\alpha \in [0, 1]$ controls the relative weight of macro-induced similarity ($\mathbf{A}_{\mathrm{macro}}$) versus microbial co-occurrence ($\mathbf{A}_{\mathrm{co}}$) in the combined adjacency matrix $\mathbf{A} = \alpha \mathbf{A}_{\mathrm{macro}} + (1 - \alpha) \mathbf{A}_{\mathrm{co}}$. To investigate its effect, we perform a full grid search at each $\alpha \in \{0, 0.1, 0.2, \dots, 1.0\}$ and record the best LOOCV accuracy achievable at each value (Figure~\ref{fig:weights}(a)).

% \begin{wrapfigure}{r}{7cm}
% \centering
% \includegraphics[width=1\linewidth]{figures/fig_alpha.pdf}
% \caption{Sensitivity to the graph mixing parameter $\alpha$. The shaded region marks the optimal range $\alpha \in [0.1, 0.9]$ where both graph components actively contribute (Accuracy = 0.8462, Macro-F1 = 0.8250). Endpoints $\alpha{=}0$ (only $\mathbf{A}_\mathrm{co}$) and $\alpha{=}1$ (only $\mathbf{A}_\mathrm{macro}$) yield noticeably lower scores.}
% \label{fig:alpha}
% \end{wrapfigure}

\noindent \textbf{Both graph sources jointly support optimal performance.}
The best accuracy of 84.62\% is achieved across a wide plateau $\alpha \in [0.1, 0.9]$, showing that the model is robust to the mixing ratio as long as both sources contribute. At the extremes, removing $\mathbf{A}_\mathrm{macro}$ ($\alpha{=}0$) drops accuracy to 69.23\%, while relying solely on $\mathbf{A}_\mathrm{macro}$ ($\alpha{=}1$) gives 76.92\%, confirming that neither source alone matches the dual-source design.
This broad plateau is important because it suggests that the gain of GRMLR is not tied to a fragile hyperparameter setting. Even when the balance between the two graph sources changes substantially, the model remains in the same high-performance region as long as both kinds of ecological information are preserved. In practical terms, this means that the framework depends more on the existence of complementary structure than on a tuned value of $\alpha$, which is a desirable property under extreme small-sample conditions.

\noindent \textbf{Macro-induced structure is dominant but not self-sufficient.}
The drop at $\alpha = 0$ (co-occurrence only) equals that at $\alpha = 1$ (macrofauna only), yet the optimal point is $\alpha = 0.1$, where co-occurrence edges carry nine times more weight. This suggests that macrofauna correlations supply the core ecological structure, while the denser co-occurrence edges fill connectivity gaps needed for effective Laplacian smoothing.

\subsection{Interpretability Analysis (RQ4)}

A key property of GRMLR is its \emph{intrinsic interpretability}: the weight matrix $W \in \mathbb{R}^{K \times p}$ directly reflects each taxon's contribution to classifying each developmental stage. We examine feature importance from two angles.

\begin{figure}[t]
\centering
\includegraphics[width=0.95\linewidth]{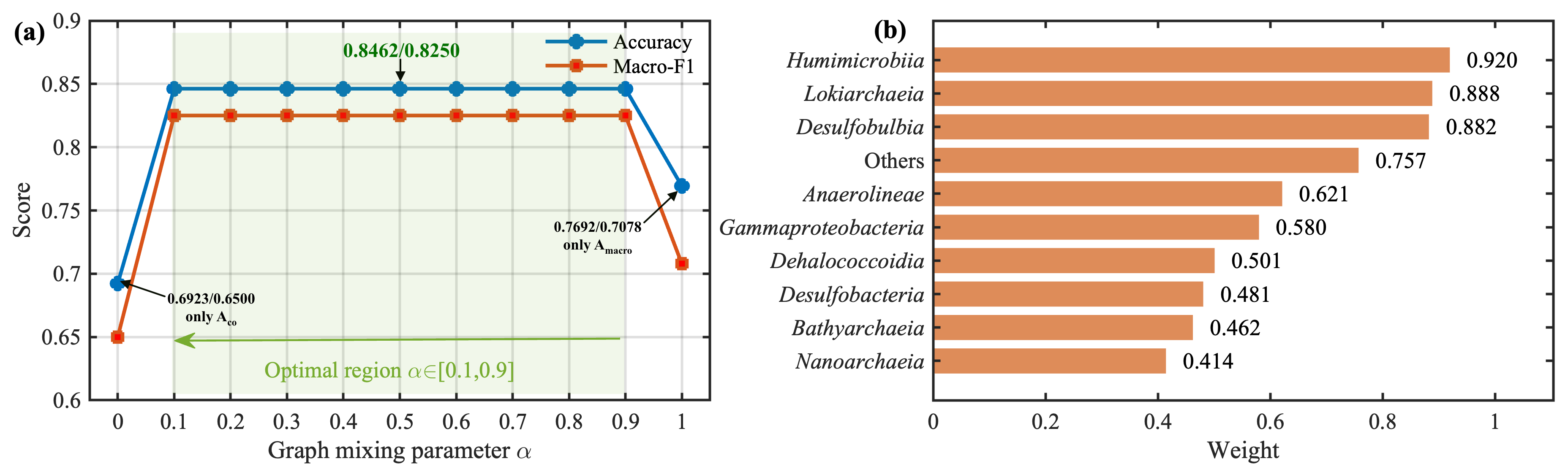}
\caption{(a) Sensitivity to the graph mixing parameter $\alpha$. The shaded region marks theoptimal range $\alpha \in [0.1, 0.9]$ where both graph components actively contribute. (b) Top-10 microbial taxa by GRMLR coefficient magnitude.}
\label{fig:weights}
\end{figure}

\noindent\textbf{GRMLR Coefficient Magnitude.}
For each taxon $j$, we compute the cross-class weight magnitude $\lVert w_{:,j} \rVert_2$ averaged over all LOOCV folds. The top-10 taxa are visualised in Figure~\ref{fig:weights}(b).

% \begin{figure}[t]
% \centering
% \begin{minipage}[t]{0.44\textwidth}
% \centering
% \captionof{table}{Top-10 microbial taxa by GRMLR coefficient magnitude.}
% \label{tab:importance}
% \scriptsize
% \begin{tabular}{clc}
% \toprule
% Rank & Taxon & Weight \\
% \midrule
% 1 & Humimicrobiia & 1.486 \\
% 2 & Lokiarchaeia & 1.427 \\
% 3 & Desulfobulbia & 1.420 \\
% 4 & Others & 1.245 \\
% 5 & Anaerolineae & 0.980 \\
% 6 & Gammaproteobacteria & 0.926 \\
% 7 & Dehalococcoidia & 0.826 \\
% 8 & Desulfobacteria & 0.758 \\
% 9 & Bathyarchaeia & 0.756 \\
% 10 & Nanoarchaeia & 0.708 \\
% \bottomrule
% \end{tabular}
% \end{minipage}%
% \hfill
% \begin{minipage}[t]{0.52\textwidth}
% \centering
% \vspace{0pt}
% \includegraphics[width=\textwidth]{figures/fig_feature.pdf}
% \captionof{figure}{Top-10 microbial taxa by GRMLR weight magnitude.}
% \label{fig:feature}
% \end{minipage}
% \end{figure}

\noindent \textbf{Ecological plausibility.}
The top-ranked taxa align with established cold seep biogeochemistry. \emph{Desulfobulbia}, \emph{Desulfobacteria}, and \emph{Anaerolineae} are well-known participants in sulphate reduction and anaerobic oxidation of methane (AOM)---the defining biogeochemical process of active cold seeps \cite{suess_marine_2014}. \emph{Lokiarchaeia} (Asgard archaea) are recurrently reported in deep-sea seep sediments. Their high weight magnitudes confirm that the model has captured ecologically meaningful signals rather than statistical artefacts.

\begin{figure}[t]
\centering
\includegraphics[width=\textwidth]{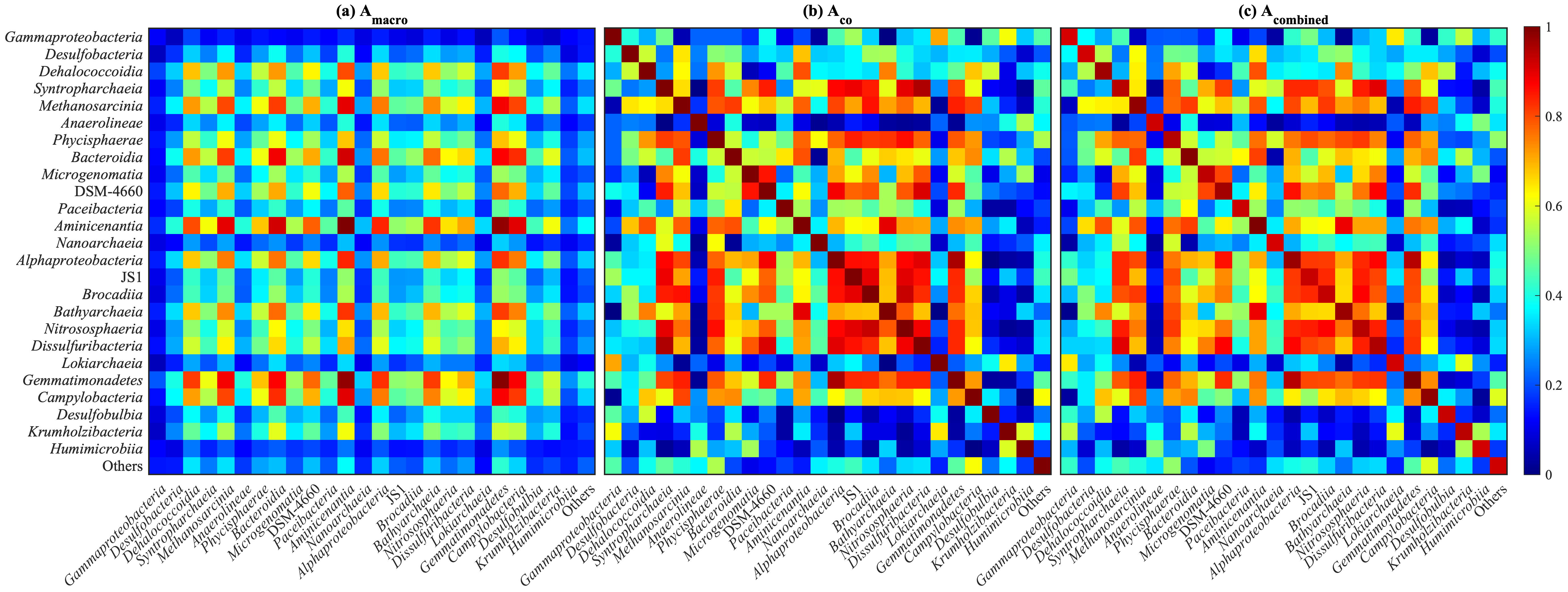}
\caption{Knowledge graph adjacency heatmaps. Left: macro-induced $\mathbf{A}_{\mathrm{macro}}$. Centre: co-occurrence $\mathbf{A}_{\mathrm{co}}$. Right: combined $\mathbf{A} = \alpha\mathbf{A}_{\mathrm{macro}} + (1-\alpha)\mathbf{A}_{\mathrm{co}}$.}
\label{fig:kg}
\end{figure}

\noindent\textbf{Knowledge Graph Visualisation.}
Figure~\ref{fig:kg} visualises the three adjacency components: $\mathbf{A}_{\mathrm{macro}}$ (macro-induced similarity), $\mathbf{A}_{\mathrm{co}}$ (co-occurrence), and the combined $\mathbf{A}$. The heatmaps show complementary structures: $\mathbf{A}_{\mathrm{macro}}$ captures sparse, ecologically grounded taxon groups that share similar responses to macrofauna, while $\mathbf{A}_{\mathrm{co}}$ adds denser connectivity from direct microbial co-occurrence. The combined graph brings both together, enabling effective Laplacian smoothing over ecologically coherent neighbourhoods.

\section{Related Work}

\noindent\textbf{Map Construction and Object Detection.} Traditional Structure-from-Motion (SfM) and Multi-View Stereo (MVS) pipelines \cite{schonberger2016structure,schonberger2016pixelwise} excel in textured scenes but frequently fail underwater due to light absorption, scattering, and uncalibrated cameras; physical restoration methods such as Sea-Thru \cite{akkaynak2019sea} mitigate image degradation yet cannot resolve geometric instability from irregular submersible motion. DUSt3R \cite{wang2024dust3r} overcomes these limitations by reformulating reconstruction as dense pointmap regression---bypassing calibration and sparse feature matching---with geometric priors from self-supervised pre-training \cite{weinzaepfel2022croco}. Our work builds upon this paradigm to achieve large-scale, consistent seafloor mapping from unconstrained deep-sea videos. In parallel, detecting macrofauna in underwater imagery remains difficult due to low contrast, scattering/backscatter, and large scale variations; recent studies increasingly build upon one-stage detectors in the YOLO family and improve robustness to degraded visual features and computational constraints via multi-scale feature extraction and lightweight designs \cite{zhai_sea_cucumber_yolov5_2022,zhao_mdg_yolo_2025}.

\noindent\textbf{Knowledge-Guided Model and Classification Model.} Parnami and Lee \cite{parnami2022learning} reviewed few-shot learning methods and showed that adding prior knowledge (for example, semantic relations and class-level prototypes) is a way to improve prediction when labeled samples are limited. Zhou et al. \cite{zhou2026knowledge} applied knowledge-enhanced pretraining to a low-sample medical classification task by injecting domain knowledge during representation learning, and reported improved diagnostic performance compared with data-driven training. These studies indicate that adding prior knowledge can stabilize learning when data are scarce.

Cortes and Vapnik \cite{cortes1995support} proposed SVM, which uses maximum-margin boundaries for supervised classification; Breiman \cite{breiman2001random} proposed random forest, which improves robustness by aggregating multiple decision trees; and Cover and Hart \cite{cover1967nearest} introduced k-nearest neighbors, which predicts labels by local neighborhood voting. For small-sample ecological data, these methods are widely used as standard baselines, but they can still overfit in high-dimensional settings. To address this issue, Belkin et al. \cite{belkin2006manifold} introduced manifold regularization, which adds graph-based smoothness constraints so related samples or features receive similar model responses, improving generalization under limited labels.

% 这里不要再介绍自己的工做了
%Our method follows the second and third lines: we encode ecological prior knowledge as a training-time graph and inject it into multinomial logistic regression via a Laplacian penalty. Concretely, the graph combines macro-induced taxa similarity and microbial co-occurrence. Different from generic graph-based classifiers, our framework is designed for train--test decoupling in cold seep assessment: macrofauna information is used only during training, and inference uses only microbial features.

\section{Conclusion}
This study presents a knowledge-graph-regularized framework to classify cold seep developmental stages using microbial relative abundance data. By integrating a learnable Knowledge Graph (KG) as a structural prior, our model effectively addresses the challenges of extreme small-sample regimes ($n=13$) and high-dimensional features ($p=26$). Experimental results on a real-world deep-sea dataset demonstrate that the proposed model achieves an accuracy of $84.62\%$, significantly outperforming standard baselines by over 22 percentage points. Our ablation studies confirm that the Knowledge Graph penalty is the most critical component, as removing this biological prior leads to a substantial $15.4$-percentage-point drop in accuracy. This highlights the KG's ability to "inject logic" and prevent overfitting when physical samples are scarce. Furthermore, the model's top predictors, such as Lokiarchaeia and Desulfobulbia, align with established ecological knowledge of methane oxidation, proving the framework's interpretability. By shifting the assessment paradigm from high-cost visual surveys to knowledge-enhanced microbial inference, this work offers a safer, more economical, and scalable solution for deep-sea resource exploration.

\bibliographystyle{splncs04}
\bibliography{references}

\end{document}